\documentclass[runningheads]{llncs}
\usepackage{graphicx}
\usepackage{amsmath}
\usepackage{amssymb}
\usepackage{algorithm2e} % typeset algorithms
\usepackage{verbatim} % for multiline comments
\usepackage[disable]{todonotes}
\usepackage{subcaption}
\usepackage{arydshln} % dashed line

\def\Error{\mathcal{L}_{\text{PCA}}}
\def\ActivationEq{\phi(v_0, \ldots, v_\ell) =  \frac{v_{1:\ell}}{\sqrt{\sum_{i=1}^\ell v_i^2}} \left(\exp\left(-v_0\sqrt{8/\pi}\right) + 1\right)^{-1/\ell}}
\def\henc{h_{\text{enc}}}
\def\hdec{h_{\text{dec}}}

\DeclareMathOperator*{\argmin}{arg\,min}
\DeclareMathOperator{\EX}{\mathbb{E}}  % expected value
\DeclareMathOperator{\Ball}{\mathcal{B}_\ell}

\begin{document}
\title{
Decentralized Differentially Private Segmentation with PATE
\thanks{This work was partially supported by the Wallenberg Artificial Intelligence, Autonomous Systems and Software Program (WASP) funded by the Knut and Alice Wallenberg Foundation.}
}
\titlerunning{Decentralized Differentially Private Segmentation with PATE}
\author{%Anonymous Authors
Dominik Fay \inst{1,2} \and
Jens Sjölund \inst{2} \and
Tobias J.\ Oechtering \inst{1}
%\inst{1}
}
\authorrunning{D.\ Fay et al.}
\institute{KTH Royal Institute of Technology, SE-100 44 Stockholm, Sweden \and
Elekta AB, Box 7593, SE-103 93 Stockholm, Sweden
}
\maketitle              % typeset the header of the contribution
\begin{abstract}

When it comes to preserving privacy in medical machine learning, two important considerations are (1) keeping data local to the institution and (2) avoiding inference of sensitive information from the trained model.
These are often addressed using federated learning and differential privacy, respectively.
However, the commonly used Federated Averaging algorithm requires a high degree of synchronization between participating institutions.
For this reason, we turn our attention to Private Aggregation of Teacher Ensembles (PATE), where all local models can be trained independently without inter-institutional communication.
The purpose of this paper is thus to explore how PATE -- originally designed for classification -- can best be adapted for semantic segmentation.
To this end, we build low-dimensional representations of segmentation masks which the student can obtain through low-sensitivity queries to the private aggregator.
On the Brain Tumor Segmentation (BraTS 2019) dataset, an Autoencoder-based PATE variant achieves a higher Dice coefficient for the same privacy guarantee than prior work based on noisy Federated Averaging.

\end{abstract}
\section{Introduction}

\begin{figure}[t]
    \centering
    \fbox{\includegraphics[width=0.45\textwidth]{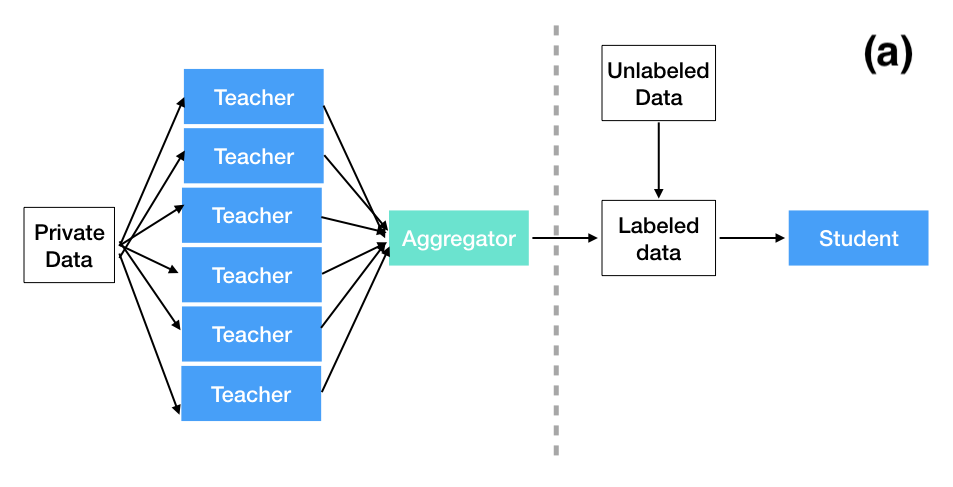}} \fbox{\includegraphics[width=0.45\textwidth]{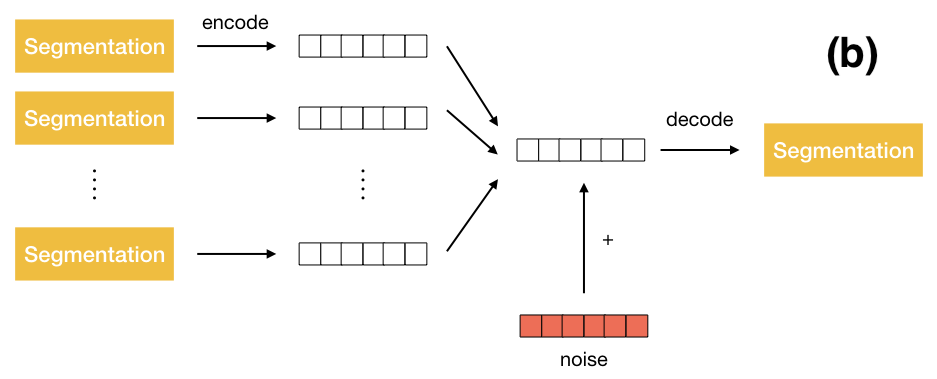}}
    \caption{Illustration of our method. \textbf{(a)} The general PATE algorithm \cite{papernot2016}. \textbf{(b)} Our proposed aggregator based on dimensionality reduction. }
    \label{fig:illustration}
\end{figure}

Image analysis methods based on deep learning are powerful but require large amounts of data. Since individual institutes may not have sufficient data on their own, they could benefit from collaborating with other institutes in order to train a joint model on shared data. This, however, presents a challenge from a privacy point of view because medical data is considered highly sensitive. The primary method of choice in the recent literature to avoid a centralized dataset has been the Federated Averaging algorithm \cite{mcmahan2016communication} and variants thereof \cite{geyer2017differentially,li2019privacy,mcmahan2018,sheller2018multi}. Here, model updates are computed locally and then averaged across institutes. The result is used to update a global model, which is then communicated back to the institutes for the next round. In a medical context, however, it may not always be possible for a large number of hospitals to engage in such a synchronized procedure at the same time.

In a different line of work, the Private Aggregation of Teacher Ensembles (PATE) has been used to train differentially private neural networks with a strong privacy-utility trade-off \cite{papernot2016,papernot2018}. Here, a model is trained in isolation for every disjoint subset of the training data. Afterwards, the local models - referred to as teachers - act as an ensemble to produce labeled data for subsequent training of a global model - referred to as the student - by predicting labels of unseen data points. This circumvents the need for synchronized communication between institutions.

In the following, we present a generic extension to PATE that incorporates a step of dimensionality reduction. In particular, we analyze three classes of dimensionality reduction methods and propose aggregation schemes tailored to them. In a series of experiments on BraTS, we try to answer empirically which dimensionality reduction method is most suitable and compare the overall segmentation quality of our approach to related work. In essence, we find that our approach compares favourably to institute-level noisy Federated Averaging \cite{geyer2017differentially} but still leaves a substantial gap to the non-private baseline. Autoencoders were the best candidates among the dimensionality reduction methods tested.

\section{Method}

\begin{algorithm}[t]
 \KwData{$K$ teacher models $t_1, \ldots, t_K$; $N$ unlabeled inputs $x_1, \ldots, x_N$; privacy parameters $\epsilon, \delta$; encoding and decoding functions $\henc$, $\hdec$}
 \KwResult{Student model}
 \For{n = 1 to N}{
    \For{k = 1 to K}{
        Run the teacher model $y_{nk} = t_k(x_n)$ \\
        Compress the prediction $z_{nk} = \henc(y_{nk})$
    }
    Draw $\gamma_n \sim \mathcal{N}(0, \sigma^2 I)$ with
    $\sigma = \frac{\sqrt{N} \left( \sqrt{\log \delta^{-1} + \epsilon} + \sqrt{\log \delta^{-1}} \right)}{\sqrt{2} K \epsilon}$ \\
    Aggregate and perturb $\bar{z}_n = \frac{1}{K} \sum_{k=1}^{K} z_{nk} + \gamma_n$ \\
    Recover the segmentation $\hat{y}_n = \hdec(\bar{z}_n)$
 }
 Train the student model on $\left((x_n, \hat{y}_n)\right)_{n=1..N}$
 
 \caption{PATE with dimensionality reduction}
 \label{alg:pate-dimred}
\end{algorithm}

We begin by presenting PATE in a non-classification, decentralized context. Then, we introduce the additional step of dimensionality reduction, including specific aggregation schemes for each proposed method and state why they preserve differential privacy.
\subsection{PATE}
The goal of PATE (see Fig. \ref{fig:illustration}(a)) is to generate a privacy-preserving labeled dataset (for subsequent training) through the use of intermediate teacher models. The private dataset consists of $K$ disjoint subsets. On each of these, an arbitrary model $t_k$ is trained. This training may happen locally at the institution that owns the respective fraction of data. Then, each teacher submits its predictions $(t_k(x_n))_{n=1..N}$ on a public unlabeled dataset to an aggregator. The aggregator collects and perturbs all teachers' predictions in such a way that the aggregate labels preserve differential privacy.

In the case of classification, the aggregator can perform a simple noisy majority voting, which is a low-sensitivity operation. For segmentation, however, the high dimensionality makes a low-sensitivity aggregation difficult. We propose to first find a low-dimensional dense representation $z_n = \henc(y_n)$ of the segmentation mask using an approximately invertible function $\henc$. On this representation we can perform a simple aggregation such as the arithmetic mean. By making sure the $\ell_2$-norm of the representation is bounded by 1, the sensitivity of the aggregation becomes $1/K$. After adding appropriately scaled Gaussian noise, the segmentation mask can be recovered through the corresponding reverse mapping $\hdec$:
\begin{align}
    \hat{y} = \hdec\left(\frac{1}{K} \sum_{k=1}^{K} \henc(y_{k}) + \mathcal{N}(0, \sigma^2 I)\right).
\end{align}
We will refer to $\henc$ and $\hdec$ as the encoder and decoder function, respectively.

We accumulate the privacy loss over multiple aggregations in terms of Rényi Differential Privacy \cite{mironov2017}. By choosing the Rényi order $\alpha$ optimally, we find that we need to scale the noise to
\begin{align}
    \sigma = \frac{\sqrt{N} \left( \sqrt{\log \delta^{-1} + \epsilon} + \sqrt{\log \delta^{-1}} \right)}{\sqrt{2} K \epsilon}
\end{align}
if we want to answer $N$ such queries under $(\epsilon, \delta)$-differential privacy. Algorithm \ref{alg:pate-dimred} provides a formal description of our procedure.

In the following sections, we present candidates for encoder/decoder functions based on dimensionality reduction.

\subsection{Principal Component Analysis}
Principal Component Analysis (PCA) performs dimensionality reduction by mapping data onto the directions of maximum variance. In this case, the encoder function is $\henc(y) = A_\ell y$ where the matrix $A_\ell = (a_1, \ldots, a_\ell)^T$ consists of the eigenvectors of the sample covariance matrix $\hat{\Sigma} = 1/(M-1) \sum_{i=1}^M y_i y_i^T$ that are associated with the $\ell$ largest eigenvalues $\lambda_1, \ldots, \lambda_\ell$. Since the eigenvectors form an orthonormal basis, projecting onto them cannot increase $\ell_2$-norm. Thus, if we clip and/or scale $y$ to at most unit $\ell_2$-norm then the representation $z = \henc(y)$ will have $\ell_2$-norm at most one as well. The decoding is performed as $\hdec(z) = A_\ell^T z$.

The well known characteristics \todo{which ones?} of PCA allow us to assess its performance not only experimentally but also in theory. If we denote by $A$ the full matrix of eigenvectors and by $\pi_\ell$ the projection onto the first $\ell$ components then we can express the expected squared error due to aggregation and perturbation as
\begin{align}
    \Error = \EX[|| y - A^T(\pi_\ell A y + \gamma) ||_2^2],
\end{align}
where $\pi_\ell$ is the matrix that projects onto the first $\ell$ entries and $\gamma \sim \mathcal{N}(0, \sigma^2 I)$. After multiplying from the left with $A$, we can see that the error decomposes into the removal of information due to PCA and noise:
\begin{align*}
    \EX[||(I - \pi_\ell) A y ||_2^2] + \EX[||\gamma||_2^2] = \EX\left[\sum_{j=l+1}^d (a_j y)^2\right] + \ell \sigma^2.
\end{align*}
Each term in the sum is the variance of $y$ along the respective principal component, for which $\lambda_j$ is an unbiased estimate, thus $\Error = \sum_{j=l+1}^d \lambda_j + \ell \sigma^2$. We can minimize this error by choosing the number of principal components $\ell$ such that, loosely speaking, only those directions with more signal than noise are retained:
\begin{align}
    \argmin_\ell \Error = \argmin_{\ell \in \{1, \ldots, d\}} \left\{\lambda_\ell | \lambda_\ell > \sigma^2\right\}.
\end{align}

\subsubsection{Blocking} Being a linear method, PCA needs a balance between the number of data points $M$ and their dimension $d$, since at most $\min\{M, d\}$ eigenvalues will be nonzero. In medical imaging, we typically have $d \gg M$. For this reason, we divide each segmentation mask into disjoint blocks of size $d_{\text{block}} \times d_{\text{block}} \times d_{\text{block}}$ and perform PCA on these. Instead of clipping $y$ to unit norm, we clip each block to norm $\sqrt{d_{\text{block}}^3/d}$ so that the norm over all blocks of a volume will be at most one. For BraTS, we choose $d_{\text{block}} = 24$.

To conclude, using PCA for the dimensionality reduction comes with both a theoretical performance guarantee as well as an analytical criterion for choosing the optimal number of components. Problems due to high dimensionality are taken care of by blocking.

\subsection{Autoencoder}

While PCA is highly interpretable and well grounded in statistical theory, the target may not always be linearly compressible. Furthermore, the arithmetic mean of the low-dimensional representations also corresponds to a linear operation in the original space, which might not be desired. In the case of images, this can lead to blur. If we use an autoencoder instead, we can address both of these shortcomings and additionally gain more control over the sensitivity, for instance by choosing the activation function appropriately in the bottleneck layer \footnote{By bottleneck layer, we mean the layer that outputs the low-dimensional representation and separates the encoder from the decoder}.

While the commonly used tanh or logistic functions could be used to bound the representation, the bound would be imposed by an $\ell$-cube -- that is, by the max-norm -- which is an inefficient use of space when we only need to bound the $\ell_2$-norm. For this reason, we choose an activation $\phi: \mathbb{R}^{\ell + 1} \rightarrow \Ball$ which maps to the unit $\ell$-ball $\Ball = \{ x \in \mathbb{R}^\ell: ||x||_2 \leq 1 \}$ directly. Thus, the norm of the representation is bounded by design and the network can be optimized with this constraint.

Furthermore, we would prefer the activations to have approximately uniform distribution because it provides the highest capacity (entropy) for distributions over bounded support. Since the inputs are usually assumed to be approximately Normally distributed (especially when using batch normalization), we construct the activation function $\phi$ such that $\phi(X) \overset{approx.}{\sim } \text{Uniform}(\Ball)$ if $X \sim \mathcal{N}(0, I)$. We do so by using the first input to determine the distance from the origin and normalizing the remaining inputs by their $\ell_2$-norm. A similar activation function has been described previously in the context of spherical regression \cite{liao2019}.
The distance from the origin is calculated by means of a scaled logistic function (to approximate the standard Normal distribution function) taken to the $\ell$-th root, the rationale being that in a uniform distribution over the unit $\ell$-ball, the distance of a randomly chosen point from the origin is distributed according to $U^{1/\ell}$ where $U \sim \text{Uniform[0, 1]}$. 
In summary, we have
\begin{align} \label{eq:activation}
    \ActivationEq.
\end{align}

\begin{figure}[t]
    \centering
    \includegraphics[width=0.6\linewidth]{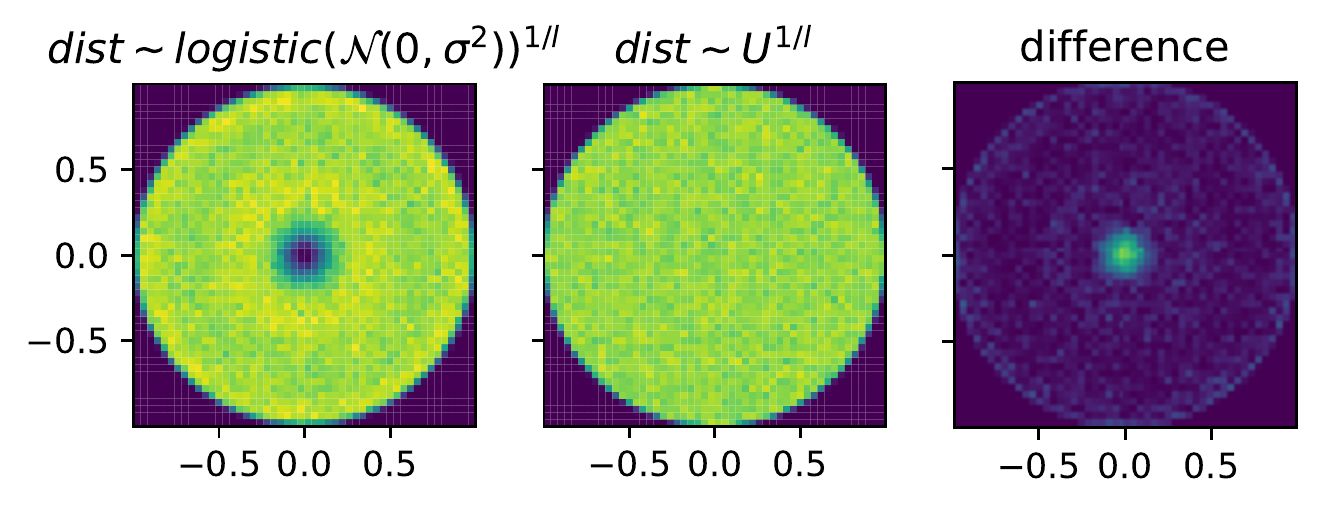}
    \caption{Samples from the distribution of activations ($\ell=2$) for standard Normal inputs (left) compared to a true uniform distribution (center) and their absolute difference (right).}
    \label{fig:activations}
\end{figure}

As Figure \ref{fig:activations} shows, the distribution of activations for standard Normal inputs for the two-dimensional case is indeed close to uniform.

\subsection{Wavelet transform}
We compare the two aforementioned dimensionality reduction methods with the Discrete Wavelet Transform (DWT)\todo{Citation?}. In contrast to PCA and Autoencoders, DWT is not learned from data. Rather, it can be seen as a general-purpose extractor of sparse representations. The transform can be made more application-specific via the choice of suitable filters. While we cannot expect the same compression rate as from learned representations, the DWT has the advantage of not requiring an additional dataset of segmentation masks.

DWT can be used to obtain a compressed representation by setting coefficients with small absolute values to zero. However, in contrast to PCA, we do not know in advance which coefficients will remain, which prevents a direct application of the Gaussian mechanism. Instead, we use the Sparse Vector Technique \cite{hardt2010multiplicative} with noise calibrated to the chosen $(\epsilon, \delta)$ to find the largest Wavelet coefficients. This is a slight deviation from the procedure shown in Alg. \ref{alg:pate-dimred} as the addition of noise is already part of the encoding.

\section{Experiments}
We perform a series of experiments on the BraTS 2019 \cite{bakas2017advancing} dataset, which consists of preprocessed multi-modal magnetic resonance imaging (MRI) brain scans from 335 subjects, manually labeled with segmentation masks corresponding to the presence of gliomas. The dataset distinguishes between three different regions of the tumor. For simplicity, we consider the binary version of the segmentation task in our experiments, that is, distinguishing the whole tumor from background.

\subsubsection{Autoencoder} The architecture for the autoencoder is a 3D fully convolutional network. The encoder part consists of 3x3x3 convolutional layers with ReLU activations, followed by max-pooling. The bottleneck layer uses 1x1x1 convolutions with the activation function described in Eq.\ \eqref{eq:activation}. The decoder part consists of convolutions with ReLU activations, followed by upsampling. \todo{More details in supplementary material?} The output layer uses sigmoid activations. Cross-entropy is used as the loss function. Gaussian noise is added to the bottleneck activations during the training process in order for the decoder to train on the same distribution as it will perform its predictions on.

\subsection{Compression, distortion and noise tolerance}
\begin{figure}[t]
    \centering
    \begin{subfigure}{0.4\textwidth}
        \centering
        \includegraphics[width=\textwidth]{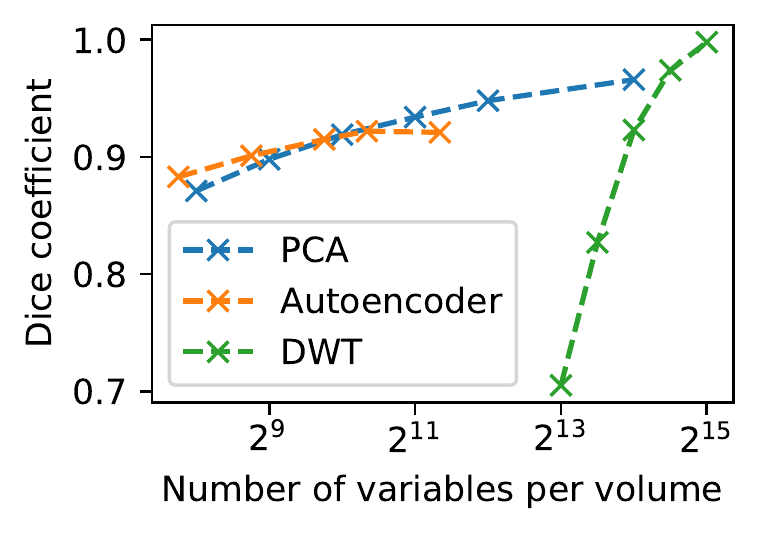}
        \label{fig:dice-dimred-a}
    \end{subfigure}
    \begin{subfigure}{0.4\textwidth}
        \centering
        \includegraphics[width=\textwidth]{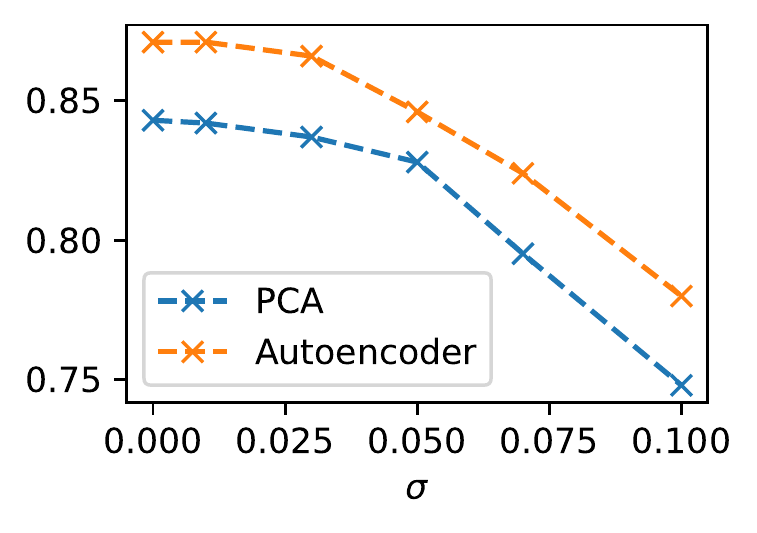}
        \label{fig:dice-dimred-b}
    \end{subfigure}
    \caption{Dice coefficient measures the utility remaining after dimensionality reduction.
    Left: Reconstruction quality depending on the number of variables per volume. 
    Right: Reconstruction quality depending on noise level.}
    \label{fig:dice-dimred}
\end{figure}

Before turning to the segmentation task, we evaluate the three dimensionality reduction algorithms separately on the set of segmentation masks. PCA and Autoencoder are trained on 100 segmentation masks and then tested on 25 different masks. Figure \ref{fig:dice-dimred} shows the performance of the three algorithms depending on the number of dimensions in their representation and noise level. For low compression rates, we notice that DWT provides superior performance but drops off steeply as coefficients get set to zero. For strong compression rates, Autoencoders perform best. Both PCA and Autoencoders behave reasonably well under the addition of noise.

Since the high-compression region is our primary interest, we use Autoencoders for the segmentation experiments which are presented in the following section.

\subsection{Segmentation}

\begin{figure}[t]
\centering
\begin{minipage}{0.4\textwidth}
    \centering
    \begin{tabular}{l|c}
        Actor & Dice \\
        \hline
        Teacher & 0.802 \\
        Teacher ensemble & 0.811 \\
        Teacher ensemble + noise & 0.799 \\
        Student & 0.785 \\
        \hdashline
        Private baseline & 0.756 \\
        Non-private baseline & 0.869
    \end{tabular}
    \captionof{table}{Segmentation performance at various stages of the training process. Baselines are shown below the dashed line.}
    \label{tab:actor-dice}
\end{minipage}
\begin{minipage}{0.2\textwidth}\end{minipage}
\begin{minipage}{0.4\textwidth}
    \centering
    \includegraphics[width=\linewidth]{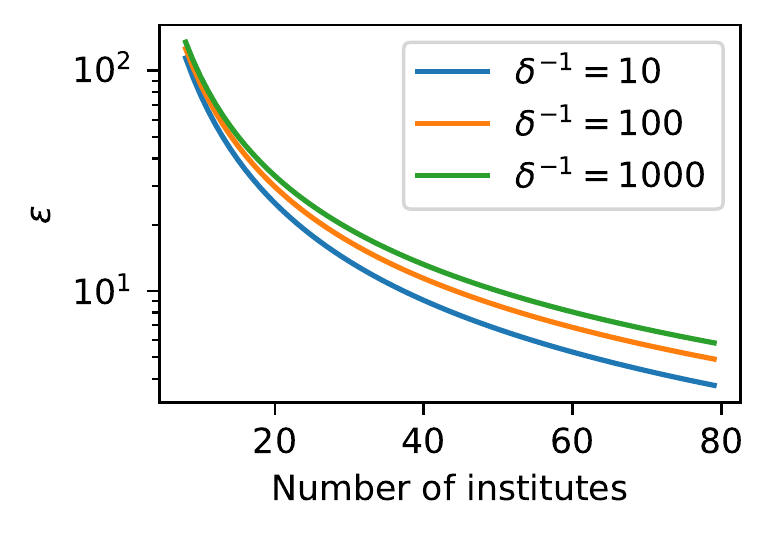}
    \caption{Level of differential privacy for $\sigma=0.075$ depending on the number of teachers.}
    \label{fig:K-eps}
\end{minipage}
\end{figure}

In order to evaluate segmentation performance, we split the dataset into $K=8$ teacher partitions of size $31$, one student partition of size $62$ and a test set of size $25$. After teacher training, the volumes contained in the student partition are labeled by the teacher ensemble under a noise level of $\sigma=0.075$ and the student is trained accordingly. The segmentation masks contained in the student partition are used to train the Autoencoder. This leads to a privacy cost of $\epsilon = 125.94$ and $\delta = 10^{-2}$. Performance on the test set is reported in Table \ref{tab:actor-dice}. Figure \ref{fig:K-eps} shows the level of privacy we would have obtained if there had been more partitions, all other things being equal. For a single-digit $\epsilon$, 37 partitions would have been necessary.

For the sake of comparability, all segmentation networks in our experiments (students, teachers, baselines) have the same architecture. We choose the well known 2D U-Net \cite{ronneberger2015u} because our primary interest is not necessarily in maximizing performance but in investigating the effect of dimensionality reduction. Our model differs from the original in that we use padding with each convolution and batch normalization after each convolution layer. We use Adam with learning rate $\eta = 0.0001$.

In order to quantify the cost of privacy, we train the base model mentioned above on the full (centralized) training set (310 subjects). We use a higher learning rate ($\eta = 0.0005$) than for teachers and students. We report this as the non-private baseline in Table \ref{tab:actor-dice}.

Since noisy Federated Averaging is the main point of comparison for our decentralized PATE variant, we apply it to our base model and report the test Dice score in Table \ref{tab:actor-dice}. We iterate federated rounds until the same $(\epsilon, \delta)$-privacy as provided by our student model is reached. In particular, we use the algorithm of \cite{geyer2017differentially} since they, too, guarantee differential privacy at the institute-level.

\section{Conclusion}
We have explored the use of dimensionality reduction to answer high-dimensional queries in the context of PATE. In the case of PCA, the error can be described analytically and the number of principal components can be chosen optimally in terms of the mean squared error. For autoencoders, we have presented a suitable architecture and activation function for the bottleneck layer that can use the $\ell_2$-bounded space efficiently. Experimentally, we have seen that Autoencoders are most suitable for the BraTS dataset. This variant of PATE can achieve higher segmentation quality than that of noisy Federated Averaging with institute-level privacy. Nevertheless, the gap to the non-private baseline is still substantial (0.785 vs.\ 0.869) despite a high value for $\epsilon$. A good value for $\epsilon$ (i.e.\ single-digit) can be reached when a moderate number ($K\geq37$) of institutes participate.

\bibliographystyle{splncs04}
\bibliography{references}
\end{document}